\documentclass[acmsmall,authorversion]{acmart}
\usepackage{amsmath}
\usepackage{mathtools}
\usepackage{stmaryrd}
\usepackage{graphicx}
\usepackage{algorithmic}
\usepackage{array}
\usepackage{multirow}
\usepackage{url}
\usepackage{hyperref}
\usepackage{color}
\usepackage{colortbl}
\usepackage{framed}
\usepackage{subcaption}
\usepackage{balance}
\usepackage{booktabs}
\usepackage{bm}
\usepackage{bbm}
\usepackage{xspace}
\usepackage{lipsum}
\usepackage{enumitem}
\usepackage{xcolor}
\usepackage{circledsteps}

\definecolor{blond}{rgb}{0.98, 0.94, 0.75}

\def \ie {\emph{i.e.}}
\def \eg {\emph{e.g.}}
\def \etal {\emph{et al.}}

\newcommand{\tit}[1]{\smallbreak\noindent\textbf{#1.}}
\newcommand{\tinytit}[1]{\noindent\textbf{#1.}}

\newcommand{\oursself}{RA-T$^\mathcal{S}$\xspace}
\newcommand{\ourscross}{RA-T$^\mathcal{X}$\xspace}

\newcommand{\rev}[1]{\textcolor{black}{#1}}

\newcommand{\selfatt}{\mathcal{S}}
\newcommand{\crossatt}{\mathcal{C}}

\AtBeginDocument{%
  }

\setcopyright{acmlicensed}
\acmJournal{TOMM}
\copyrightyear{2024}
\acmYear{2024}
\acmJournal{TOMM}
\begin{document}

%%
%% The "title" command has an optional parameter,
%% allowing the author to define a "short title" to be used in page headers.
\title[Towards Retrieval-Augmented Architectures for Image Captioning]{Towards Retrieval-Augmented Architectures for\\Image Captioning}

%%
%% The "author" command and its associated commands are used to define
%% the authors and their affiliations.
%% Of note is the shared affiliation of the first two authors, and the
%% "authornote" and "authornotemark" commands
%% used to denote shared contribution to the research.
\author{Sara Sarto}
% \authornotemark[1]
\orcid{0000-0003-1057-3374}
\affiliation{%
  \institution{University of Modena and Reggio Emilia}
  % \streetaddress{}
  \city{Modena}
  % \state{}
  \country{Italy}
  % \postcode{}
}
\email{sara.sarto@unimore.it}

\author{Marcella Cornia}
\orcid{0000-0001-9640-9385}
\affiliation{%
  \institution{University of Modena and Reggio Emilia}
  % \streetaddress{}
  \city{Modena}
  % \state{}
  \country{Italy}
  % \postcode{}
}
\email{marcella.cornia@unimore.it}

\author{Lorenzo Baraldi}
\orcid{0000-0001-5125-4957}
\affiliation{%
  \institution{University of Modena and Reggio Emilia}
  % \streetaddress{}
  \city{Modena}
  % \state{}
  \country{Italy}
  % \postcode{}
}
\email{lorenzo.baraldi@unimore.it}

\author{Alessandro Nicolosi}
% \orcid{}
\affiliation{%
  \institution{Leonardo S.p.A.}
  % \streetaddress{}
  \city{Rome}
  % \state{}
  \country{Italy}
  % \postcode{}
}
\email{alessandro.nicolosi@leonardo.com}

\author{Rita Cucchiara}
\orcid{0000-0002-2239-283X}
\affiliation{%
  \institution{University of Modena and Reggio Emilia}
  % \streetaddress{}
  \city{Modena}
  % \state{}
  \country{Italy}
  % \postcode{}
}
\affiliation{%
  \institution{IIT-CNR}
  % \streetaddress{}
  \city{Pisa}
  % \state{}
  \country{Italy}
  % \postcode{}
}
\email{rita.cucchiara@unimore.it}

%%
%% By default, the full list of authors will be used in the page
%% headers. Often, this list is too long, and will overlap
%% other information printed in the page headers. This command allows
%% the author to define a more concise list
%% of authors' names for this purpose.
\renewcommand{\shortauthors}{S. Sarto et al.}
% \renewcommand{\shorttitle}{}

%%
%% The abstract is a short summary of the work to be presented in the
%% article.
\begin{abstract}
The objective of image captioning models is to bridge the gap between the visual and linguistic modalities by generating natural language descriptions that accurately reflect the content of input images. In recent years, researchers have leveraged deep learning-based models and made advances in the extraction of visual features and the design of multimodal connections to tackle this task. This work presents a novel approach towards developing image captioning models that utilize an external \textit{k}NN memory to improve the generation process. Specifically, we propose two model variants that incorporate a knowledge retriever component that is based on visual similarities, a differentiable encoder to represent input images, and a \textit{k}NN-augmented language model to predict tokens based on contextual cues and text retrieved from the external memory. We experimentally validate our approach on COCO and nocaps datasets and demonstrate that incorporating an explicit external memory can significantly enhance the quality of captions, especially with a larger retrieval corpus. This work provides valuable insights into retrieval-augmented captioning models and opens up new avenues for improving image captioning at a larger scale.
\end{abstract}

%%
%% The code below is generated by the tool at http://dl.acm.org/ccs.cfm.
%% Please copy and paste the code instead of the example below.
%%
\begin{CCSXML}
<ccs2012>
<concept>
<concept_id>10010147.10010178.10010179.10010182</concept_id>
<concept_desc>Computing methodologies~Natural language generation</concept_desc>
<concept_significance>500</concept_significance>
</concept>
<concept>
<concept_id>10010147.10010178.10010224.10010245.10010255</concept_id>
<concept_desc>Computing methodologies~Matching</concept_desc>
<concept_significance>500</concept_significance>
</concept>
<concept>
<concept_id>10010147.10010178.10010224.10010225</concept_id>
<concept_desc>Computing methodologies~Computer vision tasks</concept_desc>
<concept_significance>300</concept_significance>
</concept>
</ccs2012>
\end{CCSXML}

\ccsdesc[500]{Computing methodologies~Natural language generation}
\ccsdesc[500]{Computing methodologies~Matching}
\ccsdesc[300]{Computing methodologies~Computer vision tasks}

%%
%% Keywords. The author(s) should pick words that accurately describe
%% the work being presented. Separate the keywords with commas.
\keywords{Image Captioning, Image Retrieval, Vision-and-Language.}

% \received{1 May 2024}
% \received[revised]{}
% \received[accepted]{}

%%
%% This command processes the author and affiliation and title
%% information and builds the first part of the formatted document.
\maketitle

\section{Introduction}
\label{sec:intro}

Because of the important role it can play in connecting vision and language in multimedia systems~\cite{karpathy2015deep,anderson2018bottom,alayrac2022flamingo}, image captioning has emerged as a fundamental task at the intersection of Computer Vision, Natural Language Processing, and Multimedia.
\rev{Image captioning architectures consist of an image encoding component and a language model that produces a coherent sentence in natural language describing the visual content of the input image. Therefore, it is important to focus on developing appropriate connections between the visual and textual modality~\cite{stefanini2022show}.} In this context, recent innovations include the usage of attentive-like architectures~\cite{anderson2018bottom,pan2020x}, the incorporation of attributes and tags~\cite{you2016image,li2020oscar}, objects~\cite{anderson2018bottom}, or scene graphs~\cite{yao2018exploring,yang2019auto,li2019know}. Despite this progress, the task still features some unique open challenges, which range from having a grounded and detailed understanding of the visual input to the selection of visual objects and semantics that are worth mentioning and their proper translation into a fluent and coherent sentence. To tackle these problems, a recent popular trend has been that of increasing the size of the model~\cite{alayrac2022flamingo,zhang2021vinvl,li2022blip}, an approach which can enhance the ability to memorize information and consistently improve the quality of image descriptions even in few or zero-shot settings. This, however, comes at the cost of increasing the number of learnable parameters and, as a result, the cost of training.

The design of large-scale language models is currently encountering similar issues, where scaling the model to increase its generation capabilities has been the dominant approach so far~\cite{zhao2023survey,caffagni2024r}. An alternative design path is that of separating \textit{language modeling} from \textit{memorization} in the architecture. While the first is an inherent ability that must be learned by the model, the second can also be tackled in a semi-parametric way by presenting the model with relevant examples from the training set, which can be exploited as suggestions during the generation phase. This can be achieved through the insertion of retrieval components~\cite{borgeaud2022improving,mialon2023augmented,caffagni2024wiki} that can allow the model to attend textual tokens or hidden states coming from an external memory, rather than relying exclusively on its own activations. This approach reduces the model memorization demands and instead delegates the task to the external memory, which is capable of handling larger-scale data and can be easily accessed through approximate nearest-neighbor searches.

\begin{figure}[t]
\centering
\includegraphics[width=0.98\columnwidth]{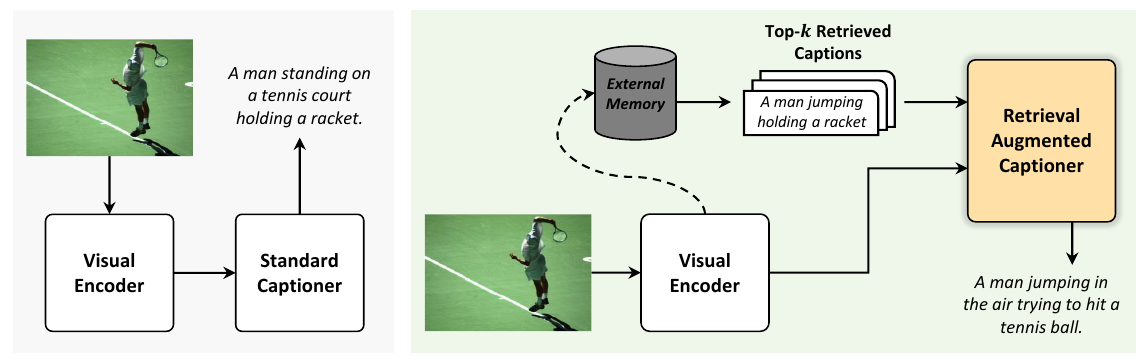}
\caption{Comparison between a standard captioner (left) and the proposed retrieval-augmented captioning schema (right), in which an external memory is employed to condition the generation process.}
\label{fig:first_page}
% \vspace{-0.2cm}
\end{figure}

In this paper, we present a thorough analysis on the design of image captioning models which follow the design path outlined above, and which are thus endowed with retrieval components (Fig~\ref{fig:first_page}). In particular, we introduce two Transformer-based architectures entirely based on fully-attentive mechanisms that integrate a knowledge retriever component and can provide the language model with appropriate cues from an external memory. The language model then exploits the information retrieved from the external memory to generate textual tokens, taking into consideration both prior context and retrieved textual elements. Following recent modeling trends, we investigate the design of an encoder-decoder protocol -- in which the connection with the external memory is based on cross-attention -- and that of a decoder-only protocol -- in which the language model is conditioned on retrieved items through self-attention. In the former case, a $k$NN-augmented attention layer is fused with the local context via a learnable gate, while in the latter case pre-processed retrieved captions are provided directly to the self-attention layers of the decoder.

Our experiments and analyses are conducted on the COCO dataset~\cite{lin2014microsoft}, the reference benchmark for standard image captioning, in comparison with Transformer-based baselines that do not leverage additional knowledge retrieved from an external memory. To assess the generalization abilities of our approach, we also validate its performance on the novel object captioning task, using the nocaps dataset~\cite{agrawal2019nocaps}. Finally, we investigate the role of the external memory content, by building different retrieval indexes with image-text pairs from both COCO~\cite{lin2014microsoft} and Conceptual Captions 3M (CC3M)~\cite{sharma2018conceptual}. The results of our experiments demonstrate that incorporating an external memory can significantly boost the generation process and improve the caption quality, especially with a larger retrieval corpus, in both of our proposed architecture variations.

\tit{Contributions} To sum up, the contributions of this paper are as follows:
\begin{itemize}[noitemsep,topsep=0pt]
    \item We propose a novel framework for image captioning in which the model is augmented with an external knowledge retrieval component. Specifically, our language model makes use of information retrieved from the external memory to generate textual tokens, considering both the previous context and the retrieved text.
    \item To validate the effectiveness of our retrieval-augmented architecture under different settings, we design two distinct fully-attentive variants, where one feeds the retrieved captions to self-attention layers while the other combines a $k$NN-augmented attention layer with the local context through a learnable gate and cross-attention.
    \item We carefully evaluate our model on COCO and nocaps, demonstrating that incorporating an external memory can effectively improve the generation process.
    \item Finally, we show that using a richer and larger retrieval index, such as one containing textual elements from CC3M, can further improve the quality and semantic richness of generated captions. 
\end{itemize}
\smallskip
This work is an improved and extended version of our conference paper~\cite{sarto2022retrieval}. With respect to this previous work, the proposed solution is extended by introducing a decoder-only architecture in which the retrieved captions are pre-processed and concatenated to the input, rather than utilizing them in cross-attention. Moreover, additional analyses and experiments are conducted, also considering a larger retrieval corpus and showing the effectiveness of both versions of our retrieval-augmented captioning model on novel object captioning.
\section{Related Work}
\label{sec:related}

\tinytit{Image Captioning} 
Image captioning is a wide-ranging task that has witnessed research on visual information extraction, text generation, and semantics incorporation. \rev{Before the advent of deep learning, traditional image captioning approaches were based on the generation of simple template sentences, which were later filled by the output of an object detector or an attribute predictor~\cite{socher2010connecting,yao2010i2t}.} \rev{Throughout the years, numerous deep learning-based techniques have been suggested: from early methods that primarily used CNN-based encoders~\cite{simonyan2014very,szegedy2015going,he2016deep} and RNN-based language models~\cite{hochreiter1997long,cho2014learning} to nowadays models that make use of attentive and Transformer-based architectures~\cite{vaswani2017attention}. The former are usually enhanced with additive attention mechanisms that can incorporate spatial knowledge, initially from a grid of CNN features~\cite{rennie2017self,Nogueira2020ReferencebasedMU,xu2015show,wang2018image} and then using image regions extracted with an object detector~\cite{anderson2018bottom,zhang2021exploring,wei2021integrating,jiang2021bi}. To further improve the encoding of objects and their relationships, graph convolution neural networks have been employed as well~\cite{yao2018exploring,yang2019auto}, to integrate semantic and spatial relationships between objects or to encode scene graphs.} 
Transformer-based models, instead, are employed both in the visual encoding stage~\cite{cornia2021explaining,liu2021cptr}, either applied directly to image patches~\cite{dosovitskiy2021image} or to refine features from a visual backbone, as well as in the language modeling part~\cite{herdade2019image,luo2021dual}. \rev{In this context, following the success of the Transformer model in Natural Language Processing (NLP) tasks~\cite{vaswani2017attention,devlin2018bert}, several captioning solutions based on fully-attentive mechanisms have been proposed, becoming the dominant choice in image captioning~\cite{cornia2020meshed,pan2023bottom,li2023exploring}.}
 
The emergence of Transformer-based models in this field has also led to the development of effective variations of self-attentive operators~\cite{pan2020x,herdade2019image,cornia2020meshed,huang2019attention,liu2020prophet} and early-fusion approaches~\cite{li2020oscar,hu2021scaling} based on BERT-like architectures~\cite{devlin2018bert} that merge vision and language features using a single Transformer stream. As for image encoding, a common and recent strategy involves using visual features extracted from large-scale cross-modal architectures~\cite{shen2022much,sarto2022retrieval,barraco2023little,caffagni2023synthcap}, such as CLIP~\cite{radford2021learning}.

This paper follows the prevalent path of utilizing visual features obtained from large-scale multimodal models and a fully-attentive language model, and proposes a retrieval-enhanced Transformer-based architecture.

\tit{Retrieval-Augmented Architectures}
The task of retrieval has been applied for decades in image similarity search, advancing over the years from the use of local descriptors to convolutional encoders, until nowadays solutions~\cite{dubey2021vision,el2021training,xu2023multi} based on ViT models~\cite{dosovitskiy2021image}. On the other side, as large-scale language models become bigger, they gain the ability to retain more information from their training data which leads to improved performance on a variety of downstream tasks~\cite{Radford2019LanguageMA,brown2020language}. This suggests that enhancing models with retrieval, thus fostering their memorization capabilities, may lead not only to further improvements but also to savings in model size. 

\rev{In the past, image~\cite{fruchard2023user,cui2015social,Zha2012,shen2011multimedia,liu2009tag} and video~\cite{shen2016accurate} tagging has been recognized as a successful practice to boost relevance matching for information retrieval. In fact, tagging is a mechanism for assigning a set of text labels (\eg~keywords or terms) to an image or a video, and can be treated as anchors to guide the visual-language alignment more explicitly. Some vision-and-language methods used tags as an additional input to boost the final performance~\cite{you2016image,gao2012visual,li2020oscar,cornia2022universal}. However, predicted tags may be incomplete, inconsistent, and sparse, especially when compared to sentences, longer paragraphs, or entire documents that usually contain more complete information.}
Recently, the same idea has been applied to language models, gaining significant interest~\cite{mialon2023augmented}. To integrate knowledge into a language model, this line of work retrieves, from an external memory, items that are related to the input, either from a single modality~\cite{borgeaud2022improving,khandelwal2019generalization,guu2020realm} or from multimodal documents~\cite{chen2022re,chen2022murag,hu2022reveal}, allowing the language model to exploit them and generate more accurate predictions. Approaches such as REALM~\cite{guu2020realm}, RAG~\cite{lewis2020retrieval}, and RETRO~\cite{borgeaud2022improving} integrate Wikipedia passages and other web-scale sources as external memory to benefit downstream knowledge-intensive tasks as, for example, question answering. Some works train the retrieval model via contrastive learning~\cite{karpukhin2020dense}, while others~\cite{guu2020retrieval} train a single-document retriever by concatenating each retrieved result with the query, to compute the final loss independently. A recent work~\cite{hu2022reveal}, instead, focuses on multimodal tasks and proposes to incorporate the retrieval scores directly into an attentive fusion module, allowing the gradients to be backpropagated through the retriever component.

In this paper, we draw inspiration from these lines of research and explore the integration of retrieval techniques in image captioning. In our case, gradients are not backpropagated into the external memory, which is crucial to the scalability of our method. To incorporate the corresponding $k$-most similar retrieved captions with the rest of our system, we perform \textit{k}NN searches on the extracted visual features, as described in~\cite{johnson2019billion}. In this way, the model can use the retrieval mechanism to access all the training data and is also, in principle, not constrained to the data seen during training. \rev{\textit{k}NN searches have also been applied in the model proposed in~\cite{Nogueira2020ReferencebasedMU} where retrieved captions are used exclusively during inference to determine the best caption among a set of predicted ones according to the METEOR score~\cite{banerjee2005meteor}, without actually contributing to the generation process as in our case.} Another related approach is that presented by Wu~\etal~\cite{wu2022memorizing}, where a gated attention module is introduced to attend to the internal states of a Transformer seen during past training iterations. In our approach, retrieval is employed on an external memory rather than on internal activations, eventually using a single scalar gate to effectively merge retrieved information with input words, instead of using a learned per-head parameter.

\section{Proposed Method}
\label{sec:method}

\subsection{Preliminaries}\label{sec:preliminaries}
Traditionally, image captioning models aim at modeling a distribution $p(y|I)$ over possible natural language descriptions $y$ given an input image $I$. During the training stage, the model is optimized with a time-wise language modeling objective, usually expressed with a cross-entropy loss. Given an image $I$ and a ground-truth caption $\hat{y}$, in the form of a sequence of tokens, the objective at each time-step $t$ is to predict a probability density over the token dictionary given previous ground-truth tokens $\{\hat{y}_\tau\}_{\tau < t}$~\cite{cornia2020meshed,cornia2022universal}. Depending on the particular tokenization algorithm of choice, tokens might correspond to entire words or to sub-words, and the cross-entropy loss encourages the predicted probability distribution to match the ground-truth token $\hat{y}_t$. An additional fine-tuning stage based on reinforcement learning might also be carried out -- in this case, the model is usually asked to generate an entire caption $y$ by relying on its own prediction of previous tokens. The generated caption is then usually matched with ground-truth captions to obtain a reward signal~\cite{rennie2017self}. 

\begin{figure}[t]
\centering
\includegraphics[width=0.9\columnwidth]{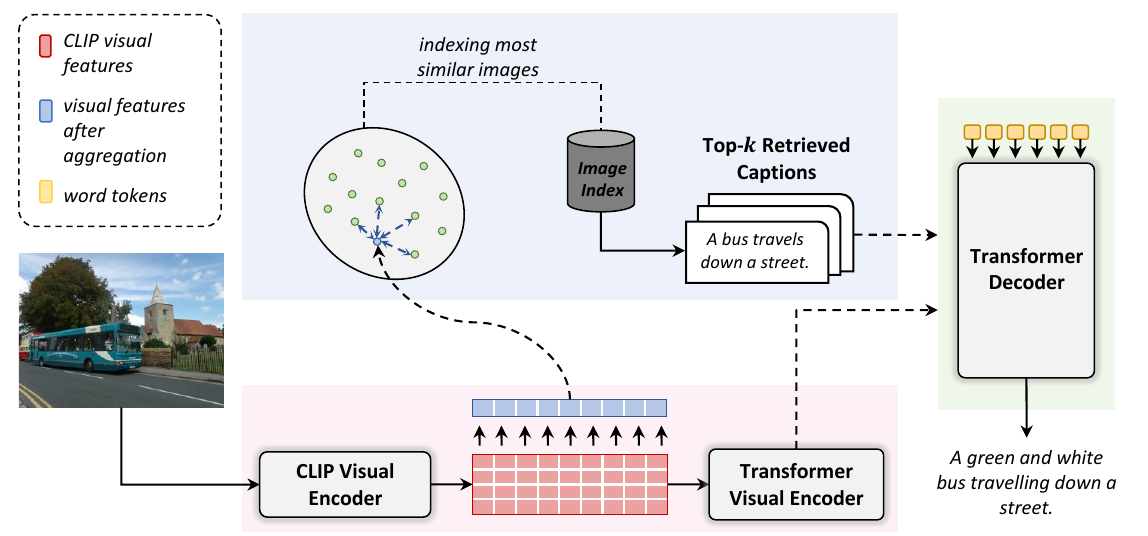}
\caption{\rev{Schema of the proposed knowledge retriever component (see Fig.~\ref{fig:architectures} for architectural details of the language models). Given an input image, visual features are extracted using a CLIP-based image encoder. These features are then used to retrieve a set of similar textual sentences, starting from the corresponding image representations, that are employed as additional knowledge during the generation of the caption.}}
\label{fig:architecture}
% \vspace{-0.2cm}
\end{figure}

With the aim of separating the language modeling and memorization capabilities of the captioner, we augment the model with an external memory of textual descriptions (Fig.~\ref{fig:architecture}), which will serve as the memory of the model. Under this setting, we can decompose the modeling of the probability distribution $p(y|I)$ into two steps: \Circled{1} \textit{retrieval} of relevant textual items from the external memory and \Circled{2} \textit{prediction} of the textual description (or language modeling), conditioned on retrieved items. Firstly, given an image $I$ we retrieve a set of descriptions $\{z_i\}_i$ from the external memory, performing $k$NN searches in a visual similarity space. Then, we condition our language model on both the input image $I$ and the set of retrieved descriptions $\{z_i\}_i$. From a probabilistic point of view, this amounts to modeling $p(y|\{z_i\}_i,I)$ and marginalizing over the set of retrieved captions.

\subsection{External Memory and Knowledge Retrieval}
\label{sec:retriever}
The retrieval of relevant textual items from the external memory aims at modeling $p(z|I)$ given a corpus of image-text pairs and an input query image $I$. This is done by performing an approximate nearest-neighbor search in the external memory, which we define through an inner product similarity between image embeddings. The relevance function $f(\cdot, \cdot)$ between the query image and images in the corpus is defined as
\begin{equation}
    f(I_1, I_2) = \texttt{Embed}(I_1)^\intercal \texttt{Embed}(I_2),
    \label{eq:relevance}
\end{equation}
where $\texttt{Embed}(\cdot)$ is a function that maps an image to a vector. After $k$NN lookup, the external memory can return all the captions of the selected images, which can then be employed as a source of conditioning for the language model.

To encode input images and build the $\texttt{Embed}(\cdot)$ function, we choose to employ the visual encoder of a CLIP model~\cite{radford2021learning} based on the ResNet~\cite{he2016deep} architecture. In this kind of architecture, the classical ResNet structure is completed with an attention pooling layer: here, a single learnable query is employed to perform attention over the grid of output features of the last convolutional layer, so to pool the grid of activation in an attention-aware fashion. In our case, we remove the attention pooling layer and adopt a custom aggregation function (\eg~average or max) so to have a more fine-grained visual representation and a higher control of the pooling procedure.

Compared to more traditional visual feature backbones (\eg~CNNs trained on visual classification) the CLIP backbone has the advantage of having been trained to match image-text pairs, which creates a representation that is inherently multimodal. Empirically, we observed that this representation is more robust with respect to vision-only descriptors, as also reported in recent literature~\cite{shen2022much,barraco2023little}.

\subsection{Designing Retrieval-Augmented Language Models}
Given an external memory from which a set of relevant captions $\{z_i\}_i$ can be extracted, we now discuss the design of a retrieval-augmented language model $p(y|\{z_i\}_i,I)$, which is in charge of predicting the output caption while being conditioned on both the input image and items retrieved from the external memory. Compared to a traditional image captioning model, which only models $p(y|I)$, a retrieval-augmented model must implement a connection between its inherent language modeling capabilities, which in a Transformer-based model take place in self-attention layers, and the sequences of tokens that form the retrieved captions. 
The framework we employ for caption generation is an encoder-decoder Transformer~\cite{vaswani2017attention}, where the encoder is in charge of processing the input image, while the decoder acts as language model. 

\rev{\subsection{Visual Encoder}}
The input of the encoder is a sequence of grid feature vectors extracted from the input image (see Sec.~\ref{sec:retriever}). Each encoder layer is then composed of a self-attention layer and a feed-forward layer, as in the standard Transformer~\cite{vaswani2017attention}. 

\rev{In particular, all intra-modality interactions between image-level features are modeled via scaled dot-product attention, without using recurrence. Attention operates on three sets of vectors, namely a set of queries $\bm{Q}$, keys $\bm{K}$ and values $\bm{V}$, and takes a weighted sum of value vectors according to a similarity distribution between query and key vectors. In the case of scaled dot-product attention, the operator can be formally defined as
\begin{align}
\mathsf{Attention}(\bm{Q}, \bm{K}, \bm{V})=\operatorname{softmax}\left(\frac{\bm{Q} \bm{K}^{T}}{\sqrt{d}}\right) \bm{V},
\label{eq:attention}
\end{align}
where $\bm{Q}$ is a matrix of $n_q$ query vectors, $\bm{K}$ and $\bm{V}$ both contain $n_k$ keys and values, all with the same dimensionality, and $d$ is a scaling factor.}

\rev{Given a set of grid image features $\bm{X}$ extracted from an input image, attention is used to obtain a permutation invariant encoding of $\bm{X}$ through the self-attention operations used in the Transformer~\cite{vaswani2017attention}. In this case, queries, keys, and values are obtained by linearly projecting the input features, and the operator can be defined as 
\begin{equation}
    \selfatt(\bm{X}) = \mathsf{Attention}(W_q\bm{X}, W_k\bm{X}, W_v\bm{X}),
\end{equation}
where $W_q, W_k, W_v$ are matrices of learnable weights. 
The output of the self-attention operator is a new set of elements $\selfatt(\bm{X})$, with the same cardinality as $\bm{X}$, in which each element of $\bm{X}$ is replaced with a weighted sum of the values, \ie~of linear projections of the input (Eq.~\ref{eq:attention}). }

\rev{The output of the self-attention attention is then applied to a position-wise feed-forward layer composed of two affine transformations with a single non-linearity, which are independently applied to each element of the set. Each of these sub-components (self-attention and position-wise feed-forward) is then encapsulated within a residual connection and a layer norm operation. The complete definition of an encoding layer can be finally written as:
\begin{align}
    \bm{J} &= \mathsf{AddNorm}(\selfatt(\bm{X})) \nonumber \\
    \bm{\tilde{X}} &= \mathsf{AddNorm}(\mathcal{F}(\bm{J})),
\end{align}}
\rev{where $\mathcal{F}$ indicates a feed-forward layer and $\mathsf{AddNorm}$ indicates the composition of a residual connection and of a layer normalization.}

\rev{Given the aforementioned structure, multiple encoding layers are stacked in sequence, so that the $i$-th layer consumes the output set computed by layer $i-1$. This amounts to creating multi-level encodings of the relationships between image features, in which higher encoding layers can exploit and refine relationships already identified by previous layers. A stack of $N$ encoding layers will therefore produce an output $\mathcal{\tilde{X}} = \bm{\tilde{X}}^N$, obtained from the output of the last encoding layer.}

\rev{\subsection{Textual Decoder}}
The decoder, instead, takes as input the sequence of tokens comprising the ground-truth caption and is asked to predict a left-shifted version of it. The self-attention here is masked so that each token can attend only elements to its left, and the decoder then effectively models an autoregressive generation process. Each layer of the decoder comprises at least one self-attention layer, a cross-attention layer with the encoder output, and one feed-forward layer.

\rev{Given an input sequence of vectors $\bm{Y}$, and outputs from the last encoder layer $\mathcal{\tilde{X}}$, the cross-attention operators connects $\bm{Y}$ to all elements in $\mathcal{\tilde{X}}$. Formally, the operator is computed using queries from the decoder and keys and values from the encoder:
\begin{equation}
        \crossatt(\bm{\tilde{X}}^i, \bm{Y}) =  \mathsf{Attention}(W_q\bm{Y}, W_k\bm{\tilde{X}}^i, W_v\bm{\tilde{X}}^i).
\end{equation}}

\rev{As the prediction of a word should only depend on previously predicted words, the decoder layer comprises a masked self-attention operation that connects queries derived from the $t$-th element of its input sequence $\bm{Y}$ with keys and values obtained from the left-hand subsequence, \ie~$\bm{Y}_{\le t}$. Also, the decoder layer contains a position-wise feed-forward layer, and all components are encapsulated within $\mathsf{AddNorm}$ operations. The final structure of the decoder layer can be written as:
\begin{align}
    \bm{J} &= \mathsf{AddNorm}(\crossatt(\mathcal{\tilde{X}}, \mathsf{AddNorm}(\selfatt_{\text{mask}}(\bm{Y}))) \nonumber \\
    \bm{\tilde{Y}} &= \mathsf{AddNorm}(\mathcal{F}(\bm{J})),
\end{align}
where $\bm{Y}$ is the input sequence of vectors and $\selfatt_{\text{mask}}$ indicates a masked self-attention over time.
Finally, our decoder stacks together multiple decoder layers, helping to refine both the understanding of the textual input and the generation of the next tokens.
Overall, the decoder takes as input word vectors, and the $t$-th element of its output sequence encodes the prediction of a word at time $t+1$, conditioned on $\bm{Y}_{\le t}$. After taking a linear projection and a softmax operation, this encodes a probability over tokens in the dictionary.}

\begin{figure*}[t]
\centering
% \subfloat[]
{\includegraphics[width=\columnwidth]{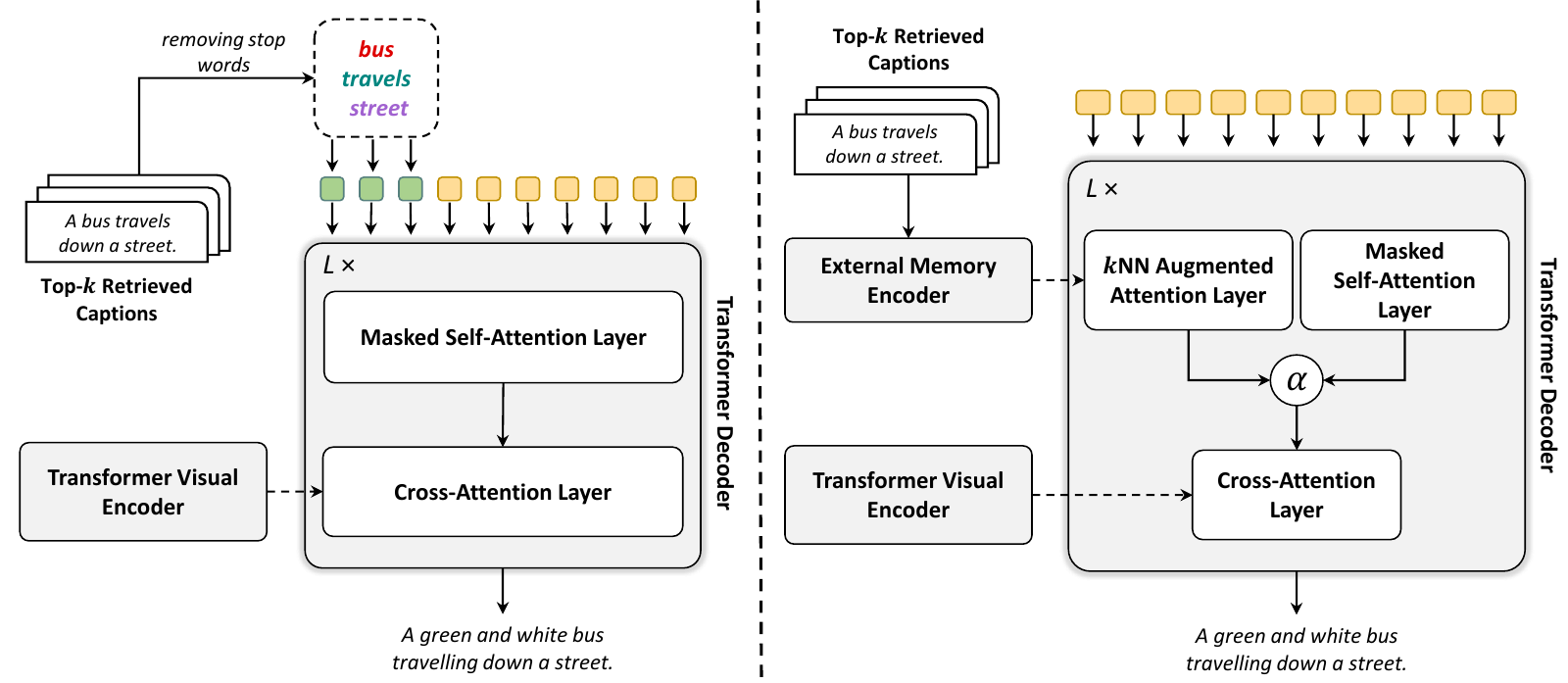}}
\caption{Architectural schema of the \oursself (self-attention-based) and \ourscross (cross-attention-based) language models. \rev{In \rev{\oursself}, retrieved captions are employed as prefix of the decoder textual sequence, after removing stop words and duplicate words. In \ourscross, instead, retrieved captions are first passed through a Transformer encoder and then used in a $k$NN cross-attention layer inside the captioner decoder. The contribution of retrieved captions is regulated by a learnable gating mechanism that combines the output of the $k$NN cross-attention layer with those of the standard self-attention over the input sequence.} }
\label{fig:architectures}
% \vspace{-0.2cm}
\end{figure*}

\rev{\subsection{Retrieval-Augmented Generation}}
As shown in Fig.~\ref{fig:architectures}, we devise two architectural variants for realizing the connection between the decoder self-attention and items retrieved from the external memory, one based on self-attentive connections, termed \oursself, and one based on cross-attention connections, termed \ourscross.

\smallskip
\subsubsection{\textbf{\oursself}} 
Under this configuration, retrieved captions are employed as a prefix of the decoder sequence, so that the self-attention operator can naturally retrieve relevant suggestions coming from the external memory while generating a caption. This might also be seen as a variant of the prompting technique~\cite{arora2022ask}.
A naïve concatenation of the retrieved captions would however be computationally intractable with the growth of $k$; furthermore, the self-attention layer would not have a principled way of distinguishing retrieved and generated tokens. Therefore, we adopt two strategies: we clean the retrieved captions by removing stop words and eliminating duplicate words that appear in more than one caption, so to obtain a set of unique words. \rev{Formally, the input of the decoder under this setting can be defined as}
\rev{
\begin{equation}
    \bm{Y}_{\text{RA-T}} = \left[\texttt{unique}(\{z_i\}_i) , \bm{Y} \right],
\end{equation}
where $\left[\cdot, \cdot\right]$ indicates concatenation, and $\texttt{unique}(\cdot)$ indicates a function that removes stop words and eliminates duplicates.}

\rev{To increase the effectiveness of this strategy, we also employ two different learnable segment embeddings~\cite{devlin2018bert} to distinguish between retrieved words and generated ones (\ie, $\bm{Y}$). Also, as cleaned words represent an unordered set, we do not apply position embeddings to this segment, so to keep the permutation invariance of the self-attention operator.}

\smallskip
\subsubsection{\textbf{\ourscross}}
In this case, an additional cross-attention layer is placed in parallel to the masked self-attention layer of the decoder. Retrieved captions are firstly encoded independently through a bidirectional Transformer encoder to get a refined representation of the tokens, then the aforementioned cross-attention layer performs a cross-attention over the resulting outputs. As the cross-attention layer is placed in parallel to the masked self-attention layer, the same queries are employed for both layers. 
\rev{Formally, given the input sequence of tokens $\bm{Y}$ and the set of retrieved captions $\bm{Z} = \{z_i\}_i$, this configuration can be written as follows:
\begin{align}
    \tilde{\bm{Z}} &= \text{Encoder}(\bm{Z}) \\
    \tilde{\bm{L}} &= \mathsf{AddNorm}(\selfatt_{\text{mask}}(\bm{Y})) \\
    \tilde{\bm{M}} &= \mathsf{AddNorm}(\crossatt(\tilde{\bm{Z}}, \bm{Y})),
\end{align}
where $\text{Encoder}$ indicates a Transformer encoder, such as the one employed for visual features encoding. The second equation refers to the self-attention between tokens of the caption. The last equation, instead, refers to the additional cross-attention operation with retrieved captions. Noticeably, in this layer, all tokens from all retrieved captions are attended.}

\rev{Finally, the outputs coming from the two parallel layers need to be combined. To this aim, we devise a learnable gate, with which the model can regulate the importance of the output coming from the self-attention layer and that coming from the cross-attention layer. Conceptually, this amounts to choosing between the local context encoding and retrieved captions. Formally,
\begin{equation}
    \tilde{\bm{J}} = \alpha \cdot \tilde{\bm{L}} + (1-\alpha) \tilde{\bm{M}},
\end{equation}
where $\alpha$ represents the learnable gate. In practice, this is learned as the sigmoid of a single scalar network parameter. The output of this linear combination is then passed to the usual cross-attention with visual features and the feed-forward network to obtain the output sequence.}

\subsection{\rev{Training Protocol}}\label{sec:training}
\rev{As briefly introduced in Sec.~\ref{sec:preliminaries}, at training time the input of the decoder is the ground-truth sentence $\{\text{BOS}, \hat{y}_1, \hat{y}_2, ..., \hat{y}_n\}$, and the model is trained with a cross-entropy loss to predict the shifted ground-truth sequence, \ie~$\{ \hat{y}_1, \hat{y}_2, ..., \hat{y}_n, \text{EOS} \}$, where $\text{BOS}$ and $\text{EOS}$ are special tokens to indicate the start and the end of the caption. While during training, the model predicts all output tokens simultaneously, the prediction process at inference happens sequentially. In each step, the model takes the partially decoded sequence as input, then selects the next token by sampling from its output probability distribution, continuing this process until an $\text{EOS}$ marker is generated.}

\rev{Following previous works~\cite{ranzato2015sequence,rennie2017self,anderson2018bottom}, after a pre-training step using cross-entropy, we further optimize the sequence generation using Reinforcement Learning. Specifically, we implement a variation of the self-critical sequence training method~\cite{rennie2017self}, which employs the REINFORCE algorithm on sequences generated through the beam search algorithm~\cite{anderson2018bottom}. Additionally, unlike the approach in~\cite{rennie2017self,anderson2018bottom}, we establish a baseline for the reward using the mean of the rewards rather than relying on greedy decoding. Specifically, given the output of the decoder, we sample the top-$k$ words from the decoder probability distribution at each timestep and always maintain the top-$k$ sequences with the highest probability. We then compute the reward of each sentence $y^i$ and backpropagate with respect to it. The final gradient expression for one sample is thus:
\begin{equation}
    \nabla_\theta L(\theta) = -\frac{1}{k}\sum_{i=1}^k \left((r(y^i)-b) \nabla_\theta \log p(y^i)\right)
\end{equation}
where $b = \left(\sum_i r(y^i)\right)/k$ is the baseline, computed as the mean of the rewards obtained by the sampled sequences. 
To reward the overall quality of the generated caption, we use the image captioning metric which better correlates with human judgment, namely CIDEr~\cite{vedantam2015cider}.}

\section{Experiments}
\label{sec:experiments}
\subsection{Experimental Setup}
\tinytit{Datasets and Evaluation Protocol} We first analyze the effectiveness of our retrieval-augmented architecture on the COCO dataset~\cite{lin2014microsoft}. \rev{We select this dataset as it is the reference evaluation benchmark for image captioning models~\cite{stefanini2022show} as well as the largest dataset with human-collected annotations for the task. In particular, COCO contains more than $120,000$ images, each of them manually annotated with five textual captions collected by using Amazon Mechanical Turk. To train and test our solution, we follow the splits defined by Karpathy~\etal~\cite{karpathy2015deep}, where the training set is composed of 82,783 images while both the validation and test set contain 5,000 images. However, there are also 30,504 images that were originally in the validation set of the original COCO dataset but were left out in this split. As done by previous image captioning~\cite{cornia2020meshed,huang2019attention,pan2020x,anderson2018bottom} and image-text matching~\cite{lee2018stacked,cheng2022cross,yao2023cross,li2023towards,messina2022aladin} literature, we add these images in the training set thus obtaining a total of 113,287 images to train the model.} 
\rev{Additionally, we perform experiments on the nocaps dataset~\cite{agrawal2019nocaps} which has been introduced for the novel object captioning task where the goal is to effectively describe objects not present in the image-caption pairs used to train the captioning model. The dataset contains 4,500 validation and 10,600 test images from the Open Images V4 dataset~\cite{kuznetsova2018open}, where each image has been annotated with 10 human-written captions.} Images can be further grouped into three subsets depending on their nearness to COCO (\ie~in-domain, near-domain, and out-of-domain images). \rev{Specifically, in-domain images only contain objects that are also present in the original COCO dataset, out-of-domain images exclusively contain object classes that are not present in COCO and thus represent the most challenging evaluation set, while near-domain images contain both in-domain and out-of-domain object classes. Under this setting, we train our model on COCO and evaluate on nocaps validation set, submitting generated captions to the nocaps evaluation server\footnote{\rev{\url{https://eval.ai/web/challenges/challenge-page/355/overview}}}.}

Following captioning literature, we measure the performance of our approach using standard captioning evaluating metrics, namely BLEU~\cite{papineni2002bleu}, METEOR~\cite{banerjee2005meteor}, ROUGE~\cite{lin2004rouge}, CIDEr~\cite{vedantam2015cider}, and SPICE~\cite{spice2016}. \rev{In particular, BLEU gauges the precision of word n-grams by comparing predicted and ground-truth sentences. As done in previous works, we evaluate our predictions with BLEU using n-grams of length 1 and 4 (referred to as B-1 and B-4, respectively). ROUGE (R) calculates an F-measure with a recall bias, using a technique based on identifying the longest common subsequence. METEOR (M), instead, assesses captions by aligning them to one or more ground-truth sentences, utilizing alignments based on various matches like exact, stem, synonym, and paraphrase between words and phrases. CIDEr (C) computes the average cosine similarity between n-grams present in the generated caption and reference sentences, employing TF-IDF weighting. SPICE (S), finally, prioritizes semantic content over fluency in generated captions by matching tuples from the candidate and reference scene graphs. Empirical evidence suggests that BLEU and ROUGE exhibit weaker correlations with human judgment compared to other metrics~\cite{vedantam2015cider,sarto2023positive}, yet the standard practice in image captioning literature involves reporting all mentioned metrics. During evaluation, we compare each generated caption with all ground-truth sentences associated with the corresponding image, using the COCO caption evaluation library\footnote{\rev{\url{https://github.com/tylin/coco-caption}}} to obtain the final scores.}

\tit{Retrieval Index} We build distinct versions of the retrieval index using image-text pairs from both COCO~\cite{lin2014microsoft} and Conceptual Captions 3M (CC3M)~\cite{sharma2018conceptual}. Specifically, when building the index on the COCO dataset, we only consider image-text pairs from the training set and, during training, we do not retrieve captions associated to the current training image to mitigate the risks of overfitting. When instead building the index on CC3M, we consider all image-text pairs available in the training set thus having an index of around 3.1M different elements. In this case, instead of using the original captions that come from noisy alt-text tags, we employ a recently proposed large-scale captioning model to produce a cleaner set of textual sentences. Specifically, we use the BLIP model~\cite{li2022blip}, in its ViT-L/14 version pre-trained on 129M image-text pairs and finetuned on the COCO dataset, to generate a new caption for each of the 3.1M images of the CC3M training set.

To improve the computational efficiency of our model, in our experiments, we utilize approximate \textit{k}NN search instead of exact \textit{k}NN search. \rev{Specifically, we use the Faiss library~\cite{johnson2019billion} and employ a graph-based Hierarchical Navigable Small Worlds (HNSW) index with 32 links per vertex.} While vector transform techniques such as PCA or vector quantization can be employed to reduce the index size and scale to larger datasets, we do not use them for simplicity.

\tit{Implementation Details}
In our model, we utilize intermediate features from CLIP-RN50$\times$16~\cite{radford2021learning} to represent images. We instead use Byte Pair Encoding (BPE)~\cite{sennrich2016neural} with a vocabulary size of $49,408$ to encode words from both the input subsequence and retrieved sentences. Standard sinusoidal positional encodings~\cite{vaswani2017attention} are employed to represent word positions. To improve efficiency, we limit the length of the output token sequence to 40 word tokens. We project visual features and word tokens into $d$-dimensional vectors with $d=384$ and pass them as input to our Transformer-based model. The model consists of $L=3$ layers in both the encoder and decoder, with six attention heads. In the \oursself version, we employ the NLTK NLP toolkit\footnote{\rev{\url{https://www.nltk.org/}}} to remove stop words. In the \ourscross, instead, the external memory encoder is composed of a single Transformer layer with the same number of heads and dimensionality as the rest of the model. The gate $\alpha$ is initialized to zero at the beginning of the training.

\tit{Training Details}
\rev{As outlined in Sec.~\ref{sec:training}, to train our solution, we employ the standard two-stage training protocol typically used by almost all captioning models:} in the first stage, we optimize the model with a time-wise cross-entropy loss, while in the second stage we perform finetuning using reinforcement learning with the CIDEr score as reward. During cross-entropy pre-training, we employ LAMB~\cite{you2019large} as optimizer with the learning rate scheduling strategy outlined in~\cite{vaswani2017attention}, using a warmup of 6,000 iterations and a batch size equal to 1,080. When instead finetuning with reinforcement learning, we use the Adam optimizer~\cite{kingma2015adam} with a batch size of 80 and a fixed learning rate equal to $5\times10^{-6}$. During both CIDEr-based finetuning and sampling of predicted captions, we employ beam search with a beam size equal to 5.

All experiments are conducted by training the models on two NVIDIA Quadro RTX-5000 GPUs, using five gradient accumulation steps for both training phases. To accelerate the training process and save memory, we use ZeRo memory offloading~\cite{rajbhandari2020zero} and mixed-precision~\cite{micikevicius2017mixed}. Overall, training the model with cross-entropy typically requires approximately 24 hours for the model that does not utilize the external memory, and around 30 hours for both variations of our complete model. Instead, finetuning with CIDEr-based optimization takes four and five days for a standard Transformer-based model and our retrieval-augmented architecture, respectively.

\begin{table*}[t]
\small
\centering
\caption{Performance of the $k$ nearest-neighbor captions using distinct retrieval indexes.}
\label{tab:simple_retrieval}
\setlength{\tabcolsep}{.35em}
\resizebox{0.85\linewidth}{!}{
\begin{tabular}{lc cccccc  c cccccc}
\toprule
& & \multicolumn{6}{c}{$k=5$}  & & \multicolumn{6}{c}{$k=10$} \\
\cmidrule{3-8} \cmidrule{10-15} 
& & B-1 & B-4 & M & R & C & S
& & B-1 & B-4 & M & R & C & S \\
\midrule
index: \textit{COCO} \\
\hspace{0.4cm}mean score 
& & 49.4 & 10.6 & 1.70 & 36.1 & 44.1 & 12.0
& & 49.2 & 10.4 & 16.8 & 35.9 & 43.1 & 11.8 \\
\hspace{0.4cm}max score (Oracle) 
& & 65.5 & 14.4 & 24.8 & 49.4 & 77.8 & 19.1
& & 72.3 & 22.1 & 28.5 & 55.1 & 96.5 & 22.6 \\
\midrule
index: \textit{CC3M (original)} \\
\hspace{0.4cm}mean score 
& & 25.5 & 0.3 & 9.3 & 20.4 & 13.5 & 5.8
& & 25.3 & 0.3 & 9.2 & 20.2 & 13.0 & 5.6 \\
\hspace{0.4cm}max score (Oracle)
& & 40.9 & 1.4 & 15.2 & 31.9 & 32.1 & 11.6 
& & 46.5 & 2.6 & 17.4 & 36.0 & 40.4 & 13.8 \\
\midrule
index: \textit{CC3M} \\
\hspace{0.4cm}mean score 
& & 59.2 & 10.7 & 20.9 & 44.2 & 61.4 & 13.8
& & 58.7 & 10.4 & 20.6 & 43.8 & 59.8 & 13.6 \\
\hspace{0.4cm}max score (Oracle)
& & 74.1 & 24.9 & 28.6 & 56.4 & 101.8 & 21.2
& & 78.5 & 31.4 & 31.3 & 60.3 & 116.6 & 23.8 \\
\midrule
\midrule
& & \multicolumn{6}{c}{$k=20$} & & \multicolumn{6}{c}{$k=40$} \\
\cmidrule{3-8} \cmidrule{10-15} 
& & B-1 & B-4 & M & R & C & S 
& & B-1 & B-4 & M & R & C & S \\
\midrule
index: \textit{COCO} \\
\hspace{0.4cm}mean score 
& & 48.9 & 10.2 & 16.7 & 35.7 & 42.1 & 11.6
& & 48.6 & 10.0 & 16.5 & 35.4 & 41.1 & 11.4\\
\hspace{0.4cm}max score (Oracle) 
& & 77.7 & 30.8 & 31.9 & 60.2 & 114.2 & 25.4
& & 82.0 & 39.4 & 34.9 & 64.5 & 130.4 & 28.0 \\
\midrule
index: \textit{CC3M (original)} \\
\hspace{0.4cm}mean score 
& & 25.2 & 0.3 & 9.1 & 20.1 & 12.6 & 5.5
& & 25.0 & 0.3 & 9.0 & 20.0 & 12.2 & 5.3 \\
\hspace{0.4cm}max score (Oracle) 
& & 51.8 & 4.2 & 19.3  & 39.9 & 49.3 & 16.1
& & 56.6 & 6.9 & 21.3 & 43.7 & 58.6 & 18.1 \\
\midrule
index: \textit{CC3M} \\
\hspace{0.4cm}mean score 
& & 58.2 & 9.9 & 20.3 & 43.4 & 58.1 & 13.3
& & 57.6 & 9.5 & 20.0 & 43.0 & 56.3 & 13.0 \\
\hspace{0.4cm}max score (Oracle)
& & 81.9 & 37.7 & 33.7 & 63.6 & 129.4 & 25.9
& & 85.1 & 43.6 & 36.0 & 66.7 & 141.8 & 27.8 \\
\bottomrule
\end{tabular}
}
\vspace{-0.15cm}
\end{table*}

\subsection{Quality of Nearest Neighbor Captions}
\rev{To confirm that nearest neighbor captions are a suitable source of additional knowledge and that can be thus employed to improve the final performance, we first need to evaluate their relevance with respect to the ground-truth captions.} To do this, given an image from the test set, we retrieve the $k$ nearest captions from one of the created nearest neighbor indexes using our relevance function to compare visual elements (\ie~in this experiment we use a standard average pooling to aggregate image features). Then, we measure the similarity between retrieved and ground-truth captions by calculating the mean captioning scores and the score of the retrieved caption with the highest similarity to the ground-truth. \rev{The latter can be considered as an upper-bound score, where an oracle evaluator is used to select the best caption among the $k$ nearest ones.} We perform this analysis using three different retrieval indexes: one containing image-text pairs from the COCO dataset and the others containing elements from the CC3M dataset, either using the original CC3M textual descriptions (\ie~\textit{CC3M (original)}) or the textual sentences predicted by the BLIP model~\cite{li2022blip}. 

The results are presented in Table~\ref{tab:simple_retrieval} as the number $k$ of retrieved sentences varies. As it can be noticed, retrieving a limited number of captions (\eg~$k=5$) leads, for all retrieval indexes, to a set of captions that only partially correlates with the ground-truth. On the other hand, increasing the number of retrieved captions slightly degrades the performance, with a decrease in the mean CIDEr score from 44.1 to 41.1 when using the index containing COCO elements. The maximum (oracle) score, instead, shows considerably higher results. Specifically, it reaches up to 130.4 and 141.8 CIDEr points, respectively using the COCO and CC3M indexes, when retrieving a large number of captions (\ie~$k=40$). \rev{The worst results, in terms of both mean and maximum scores, are obtained with the retrieval index with the original CC3M corpus which leads to 58.6 CIDEr points in terms of maximum score using $k=40$.} These results can be explained by the quality of CC3M textual sentences which are crawled from the web and, although semantically richer, have a substantially different style from the human-annotated captions contained in the COCO dataset.

Although our embedding space is built on top of state-of-the-art descriptors, the high quality achieved by the oracle captions for higher values of $k$ indicates that there is still significant room for improvement in the quality of the embedding space. \rev{It is worth noting, also, that the quality of the captions plays a crucial role in determining the quality of the embedding space, and it is not only dependent on the size of the retrieval index, as demonstrated by the results using original CC3M captions.} Therefore, even a smaller index with high-quality captions can potentially result in a better embedding space than a larger one with lower-quality descriptions.

\subsection{Model Ablation and Analysis}

\tit{Role of Different Aggregation Functions and Number of Retrieved Captions} We then examine the outcomes of various aggregation functions for aggregating visual features and retrieving the most similar images. In particular, we experiment with three different aggregation strategies: a standard average pooling over spatial features, a max pooling, and a sum of $\ell_2$-normalized features followed by an $\ell_2$-norm of the result, which has shown promising results in previous works, especially in the field of image and video retrieval~\cite{tolias2016particular}. We report the performance analysis in Table~\ref{tab:aggregation_results}, showing the results after cross-entropy pre-training. In this case, we only employ the retrieval index built on COCO and compare the results of the architecture based on self-attentive connections (\ie~\oursself) and those of the solution based on cross-attention (\ie~\ourscross). According to the results, a standard mean of grid features performs better than other aggregation functions, such as max pooling and sum of $\ell_2$-normalized features for both model variants, regardless of the number $k$ of retrieved sentences. \rev{Specifically, while $\ell_2$-norm and max-based aggregation functions respectively lead to 123.0 and 123.1 in terms of CIDEr for the \oursself models and to 123.8 and 124.0 for the \ourscross version, using a standard average pooling leads an improvement of 0.6 and 0.5 CIDEr points for \oursself and \ourscross respectively.}

\begin{table*}[t]
\small
\centering
\caption{Performance analysis of the two versions of our retrieval-augmented Transformer, by varying the number of retrieved sentences $k$ and the aggregation function used. Results are reported after cross-entropy pre-training using the COCO index.}
\label{tab:aggregation_results}
\setlength{\tabcolsep}{.45em}
\resizebox{\linewidth}{!}{
\begin{tabular}{ccc cccccc c cccccc}
\toprule
& & & \multicolumn{6}{c}{\textbf{\oursself}} & & \multicolumn{6}{c}{\textbf{\ourscross}} \\
\cmidrule{4-9} \cmidrule{11-16}
Aggregation Function & $k$ & & B-1 & B-4 & M & R & C & S & & B-1 & B-4 & M & R & C & S \\
\midrule
$\ell_2$-norm sum & 5 & & 78.2 & 38.0 & 28.5 & 58.0 & 122.2 & 21.6 & & 78.3 & 38.6 & 28.9 & 58.3 & 123.1 & 21.8 \\
$\ell_2$-norm sum & 10 & & 77.9 & 37.5 & 28.4 & 57.6 & 121.5 & 21.7 & & 78.5 & 38.6 & 28.8 & 58.3 & 122.7 & 22.1 \\
$\ell_2$-norm sum & 20 & & 77.9 & 38.0 & 28.7 & 57.9 & 123.0 & 21.7 & & 78.3 & 38.6 & 28.9 & 58.3 & 123.8 & 21.9 \\
$\ell_2$-norm sum & 40 & & 78.5 & 38.1 & 28.6 & 58.0 & 122.9 & 21.8 & & 78.2 & 39.1 & 28.7 & 57.9 & 122.8 & 22.0 \\
\midrule
max & 5 & & 78.5 & 38.2 & 28.8 & 58.3 & 122.7 & 21.8 & & 78.6 & 38.6 & 28.9 & 58.3 & 123.6 & 22.0 \\
max & 10 & & 78.2 & 38.1 & 28.4 & 58.0 & 123.1 & 21.6 & & 78.3 & 38.5 & 28.9 & 58.2 & 123.8 & 22.2 \\
max & 20 & & 78.4 & 38.3 & 28.6 & 58.4 & 123.1 & 21.7 & & 78.3 & 38.6 & 29.0 & 58.3 & 124.0 & 22.1 \\
max & 40 & & 79.0 & 38.3 & 28.6 & 58.4 & 122.9 & 21.9 & & 78.3 & 38.3 & 28.9 & 58.3 & 123.6 & 22.0 \\
\midrule
mean & 5 & & 78.7 & 38.3 & 28.7 & 58.1 & 123.3 & 22.0 & & 78.6 & 38.7 & 29.1 & 58.5 & 124.0 & 22.0 \\
\rowcolor{blond}
mean & 10 & & 78.5 & 38.4 & 28.9 & 58.1 & \textbf{123.7} & 21.9 & & 78.9 & 38.9 & 28.9 & 58.5 & \textbf{124.5} & 22.1  \\
mean & 20 & & 78.6 & 38.3 & 28.6 & 58.2 & 123.4 & 21.8 & & 78.5 & 38.6 & 28.9 & 58.3 & 124.2 & 22.0 \\
mean & 40 & & 78.4 & 37.8 & 28.7 & 58.3 & 122.9 & 21.0 & & 78.4 & 38.4 & 28.9 & 58.3 & 123.1 & 22.0 \\
\bottomrule
\end{tabular}
}
\vspace{-0.1cm}
\end{table*}

\begin{table}[t]
\small
\centering
\caption{Ablation study results of the two proposed model variants in comparison with a standard Transformer without retrieval. Results are reported after cross-entropy pre-training using the COCO index. For each model, we show the absolute improvement with respect to the baseline \rev{(\ie~Transformer (w/o external memory)}.}
\label{tab:ablation}
\setlength{\tabcolsep}{.4em}
\resizebox{0.75\linewidth}{!}{
\begin{tabular}{lccccccc}
\toprule
 & & B-1 & B-4 & M & R & C & S \\
\midrule
Transformer (w/o external memory) & & 78.1 & 38.1 & 28.5 & 58.0 & 121.6 & 21.8 \\
\textbf{\oursself} (w/ full sentences) & & 78.4 & 38.3 & 28.7 & 58.0 & 122.7 & 21.8 \\
\rowcolor{blond}
 & & \textbf{78.5} & \textbf{38.4} & \textbf{28.9} & \textbf{58.1} & \textbf{123.7} & \textbf{21.9} \\
 \rowcolor{blond}
\multirow{-2}{*}{\textbf{\oursself}} & & \textcolor{blue}{(+0.4)} & \textcolor{blue}{(+0.3)} & \textcolor{blue}{(+0.4)} & \textcolor{blue}{(+0.1)} & \textcolor{blue}{(+2.1)} & \textcolor{blue}{(+0.1)} \\
\midrule
Transformer (w/o external memory) & & 78.1 & 38.1 & 28.5 & 58.0 & 121.6 & 21.8 \\
\textbf{\ourscross} (w/o gate) & & 78.3 & 38.3 & \textbf{28.9} & 58.1 & 122.5 & 21.9 \\
\textbf{\ourscross} (w/o stop words) & & 78.7 & 38.6 & 28.8 & 58.3 & 124.0 & 22.1 \\
\rowcolor{blond}
& & \textbf{78.9} & \textbf{38.9} & \textbf{28.9} & \textbf{58.5} & \textbf{124.5} & \textbf{22.1} \\
\rowcolor{blond}
\multirow{-2}{*}{\textbf{\ourscross}} & & \textcolor{blue}{(+0.8)} & \textcolor{blue}{(+0.8)} & \textcolor{blue}{(+0.4)} & \textcolor{blue}{(+0.5)} & \textcolor{blue}{(+2.9)} & \textcolor{blue}{(+0.3)} \\
\bottomrule
\end{tabular}
}
\vspace{-0.15cm}
\end{table}

\begin{table}[t]
\small
\centering
\caption{Comparison with state-of-the-art models on the Karpathy-test split. Overall best results are underlined.}
\label{tab:sota_results}
\setlength{\tabcolsep}{.45em}
\resizebox{0.75\linewidth}{!}{
\begin{tabular}{lccccccc}
\toprule
 & & B-1 & B-4 & M & R & C & S \\
\midrule
Up-Down~\cite{anderson2018bottom} & & 79.8 & 36.3 & 27.7 & 56.9 & 120.1 & 21.4 \\
ORT~\cite{herdade2019image} & & 80.5 & 38.6 & 28.7 & 58.4 & 128.3 & 22.6 \\
GCN-LSTM~\cite{yao2018exploring} & & 80.9 & 38.3 & 28.6 & 58.5 & 128.7 & 22.1 \\
SGAE~\cite{yang2019auto} & & 81.0 & 39.0 & 28.4 & 58.9 & 129.1 & 22.2 \\ 
AoANet~\cite{huang2019attention} & & 80.2 & 38.9 & 29.2 & 58.8 & 129.8 & 22.4 \\
$\mathcal{M}^2$ Transformer~\cite{cornia2020meshed} & & 80.8 & 39.1 & 29.2 & 58.6 & 131.2 & 22.6 \\
X-LAN~\cite{pan2020x} & & 80.8 & 39.5 & 29.5 & 59.2 & 132.0 & 23.4 \\
X-Transformer~\cite{pan2020x} & & 80.9 & 39.7 & 29.5 & 59.1 & 132.8 & 23.4 \\
DPA~\cite{liu2020prophet} & & 80.3 & 40.5 & 29.6 & 59.2 & 133.4 & 23.3 \\
DLCT~\cite{luo2021dual} & & 81.4 & 39.8 & 29.5 & 59.1 & 133.8 & 23.0 \\
RSTNet~\cite{zhang2021rstnet} & & 81.8 & 40.1 & 29.8 & 59.5 & 135.6 & 23.3 \\
DIFNet~\cite{wu2022difnet} & & 81.7 & 40.0 & 29.7 & 59.4 & 136.2 & 23.2 \\
\midrule
Transformer (w/o external memory) & & 81.9 & 39.7 & 29.6 & 59.4 & 135.3 & \textbf{23.6} \\
\rowcolor{blond}
\textbf{\oursself} (index: \textit{COCO}) & & 82.0 & 40.1 &  29.6 & 59.4 &  136.4 & 23.2 \\ 
\rowcolor{blond}
\textbf{\oursself} (index: \textit{CC3M}) & & \underline{\textbf{82.5}} & \textbf{40.8} & \textbf{29.7} & \underline{\textbf{59.8}} & \underline{\textbf{136.7}} & \textbf{23.6} \\
\midrule
Transformer (w/o external memory) & & 81.9 & 39.7 & 29.6 & 59.4 & 135.3 & 23.6 \\
\rowcolor{blond}
\textbf{\ourscross} (index: \textit{COCO}) & & \textbf{82.4} & 40.5 & 29.8 & \underline{\textbf{59.8}} & 136.5 & 23.8 \\
\rowcolor{blond}
\textbf{\ourscross} (index: \textit{CC3M}) & & 82.2 & \underline{\textbf{41.0}} & \underline{\textbf{30.0}} & \underline{\textbf{59.8}} & \underline{\textbf{136.7}} & \underline{\textbf{23.9}} \\
\bottomrule
\end{tabular}
}
\vspace{-0.15cm}
\end{table}

We also evaluate the effectiveness of our retrieval strategy by varying the number of retrieved captions, using $k=5, 10, 20, 40$. Results are reported also in this case in Table~\ref{tab:aggregation_results}. As it can be noticed, using a number of retrieved captions equal to 10 and 20 generally leads to the best performance in terms of all considered evaluation metrics. Overall, the best results are obtained using the mean as aggregation function and $k=10$ retrieved captions, for both proposed model variants. Specifically, our \oursself achieves 123.7 points in terms of CIDEr, while the cross-attention-based model (\ie~\ourscross) reaches 124.5 CIDEr points. Therefore, we employ this configuration \rev{(\ie~$k=10$ and the mean as aggregation function)} for both model variations in all experimental analyses shown in the rest of the section.

\tit{Role of External Memory and Other Architectural Details} 
In Table~\ref{tab:ablation}, we first analyze the effectiveness of our retrieval-augmented solution compared to a standard encoder-decoder Transformer architecture without retrieval, with the same dimensionality and number of layers as our complete model. As it can be seen, both model configurations achieve significantly better results than the Transformer baseline, with an increase of 2.1 and 2.9 points in terms of CIDEr score respectively for \oursself and \ourscross models. These results confirm the appropriateness of employing an external memory to effectively improve the quality of generated captions.

We also evaluate the role of other architectural details as the removal of stop words from retrieved elements and the effect of the learned gate. In particular, we compare the \oursself model with a variant that takes as input the full retrieved sentences instead of removing stop words. From the results, we can notice that the model using the full retrieved captions achieves slightly worse performance, while still being better than the vanilla Transformer model. \rev{In particular, the \oursself model achieves an improvement of 1.0 CIDEr points (\ie~122.7 vs 123.7) compared to the model taking as input entire retrieved sentences, thus confirming the appropriateness of using the set of words as input for this model variant.} Regarding the \ourscross architecture, we compare it with a model without the learned gating mechanism, where masked self-attention and cross-attention between input tokens and retrieved captions are performed in sequence and a baseline that takes the same input as the \oursself counterpart (\ie~the clean set of words appearing in the retrieved items without stop words) while maintaining the structure of the \ourscross version. Also in this case, the results confirm the effectiveness of the architectural choices made, with an improvement of 2 CIDEr points (\ie~122.5 vs 124.5) and 0.5 CIDEr points (\ie~124.0 vs 124.5) respectively compared to the model without learnable gate and the baseline taking as input the captions without stop words.

\begin{table}[t]
\caption{Performances on nocaps validation set. Overall best results are underlined.}
\label{tab:nocaps}
\small
\centering
\setlength{\tabcolsep}{.4em}
\resizebox{0.75\linewidth}{!}{
\begin{tabular}{lc cc c cc c cc}
\toprule
 & & \multicolumn{2}{c}{Near} & & \multicolumn{2}{c}{Out} & & \multicolumn{2}{c}{Overall} \\
\cmidrule{3-4} \cmidrule{6-7} \cmidrule{9-10} 
& & C & S & & C & S & & C & S \\
\midrule
NBT~\cite{agrawal2019nocaps} & & 61.2 & 9.9 & & 62.4 & 8.9 & & 60.2 & 9.5 \\
Up-Down~\cite{agrawal2019nocaps} & & 73.6 & 11.3 & & 66.4 & 9.7 & & 73.1 & 11.1 \\
$\mathcal{M}^2$ Transformer~\cite{cornia2020meshed} & & 75.4 & 11.7 & & 69.4 & 10.0 & & 75.0 & 11.4 \\
\midrule
Transformer (w/o external memory) & & 87.4 & 12.7 & & 66.7 & 10.8 & & 85.3 & 12.5 \\
\rowcolor{blond}
\textbf{\oursself} (index: \textit{COCO}) & & 88.2 & 12.5 & & 68.6 & 10.6 & & 86.3 & 12.3 \\
\rowcolor{blond}
\textbf{\oursself} (index: \textit{CC3M}) & & \textbf{89.3} & \textbf{13.0} & & \underline{\textbf{69.5}} & \textbf{11.0} & & \textbf{86.8} & \textbf{12.7} \\
\midrule
Transformer (w/o external memory) & & 87.4 & 12.7 & & 66.7 & 10.8 & & 85.3 & 12.5 \\
\rowcolor{blond}
\textbf{\ourscross} (index: \textit{COCO}) & & 88.5 & 12.8 & & \textbf{68.6} & 11.0 & & 86.3 & 12.6 \\
\rowcolor{blond}
\textbf{\ourscross} (index: \textit{CC3M}) & & \underline{\textbf{89.4}} & \underline{\textbf{13.1}} & & \textbf{68.6} & \underline{\textbf{11.1}} & & \underline{\textbf{87.0}} & \underline{\textbf{12.8}} \\
\bottomrule
\end{tabular}
}
\vspace{-0.15cm}
\end{table}

\subsection{Comparison to the State of the Art}
\tinytit{Results on COCO} In Table~\ref{tab:sota_results} we report the results on the standard Karpathy test split after CIDEr-based finetuning, comparing our model performance with that of different state-of-the-art captioning models. Although several architectures pre-trained on large-scale datasets and then finetuned on COCO have recently been proposed~\cite{li2020oscar,zhang2021vinvl,hu2021scaling}, in this analysis we only consider captioning models trained exclusively on the COCO dataset. Specifically, we compare against methods with language models based on LSTMs such as Up-Down~\cite{anderson2018bottom}, eventually enhanced with spatial and scene graphs like GCN-LSTM~\cite{yao2018exploring} and SGAE~\cite{yang2019auto} or self-attentive mechanisms such as AoANet~\cite{huang2019attention}, X-LAN~\cite{pan2020x}, and DPA~\cite{liu2020prophet}. Moreover, we consider captioning architectures entirely based on the standard Transformer model such as ORT~\cite{herdade2019image}, $\mathcal{M}^2$ Transformer~\cite{cornia2020meshed}, X-Transformer~\cite{pan2020x}, and RSTNet~\cite{zhang2021rstnet}, even combining visual features from multiple backbones as in the case of DLCT~\cite{luo2021dual} and DIFNet~\cite{wu2022difnet}.

Results of both versions of our complete retrieval-augmented architecture are reported using both COCO and CC3M retrieval indexes and compared with those of a standard Transformer-based model without the retrieval component. As it can be seen, the efficacy of the $k$NN-augmented language model is confirmed even after finetuning with CIDEr-based optimization, with an increase of 1.1 and 1.2 CIDEr points respectively comparing \oursself and \ourscross with COCO retrieval index to the standard Transformer-based architecture. The use of a larger index such as the one containing a cleaned version of CC3M captions can further boost the performance, leading to an overall CIDEr score of 136.7 for both model variations. \rev{It can also be noticed that, while after cross-entropy pre-training the \ourscross version slightly outperforms the \oursself model, after reinforcement learning finetuning the two model variants perform comparably, thus demonstrating that both architectures can be a valid solution for incorporating external knowledge.} Furthermore, we observe that the proposed retrieved-augmented model achieves promising and competitive performance compared to other state-of-the-art methods, and surpasses them in terms of all evaluation metrics.

\begin{figure*}[t]
    \centering
    \includegraphics[width=\textwidth]{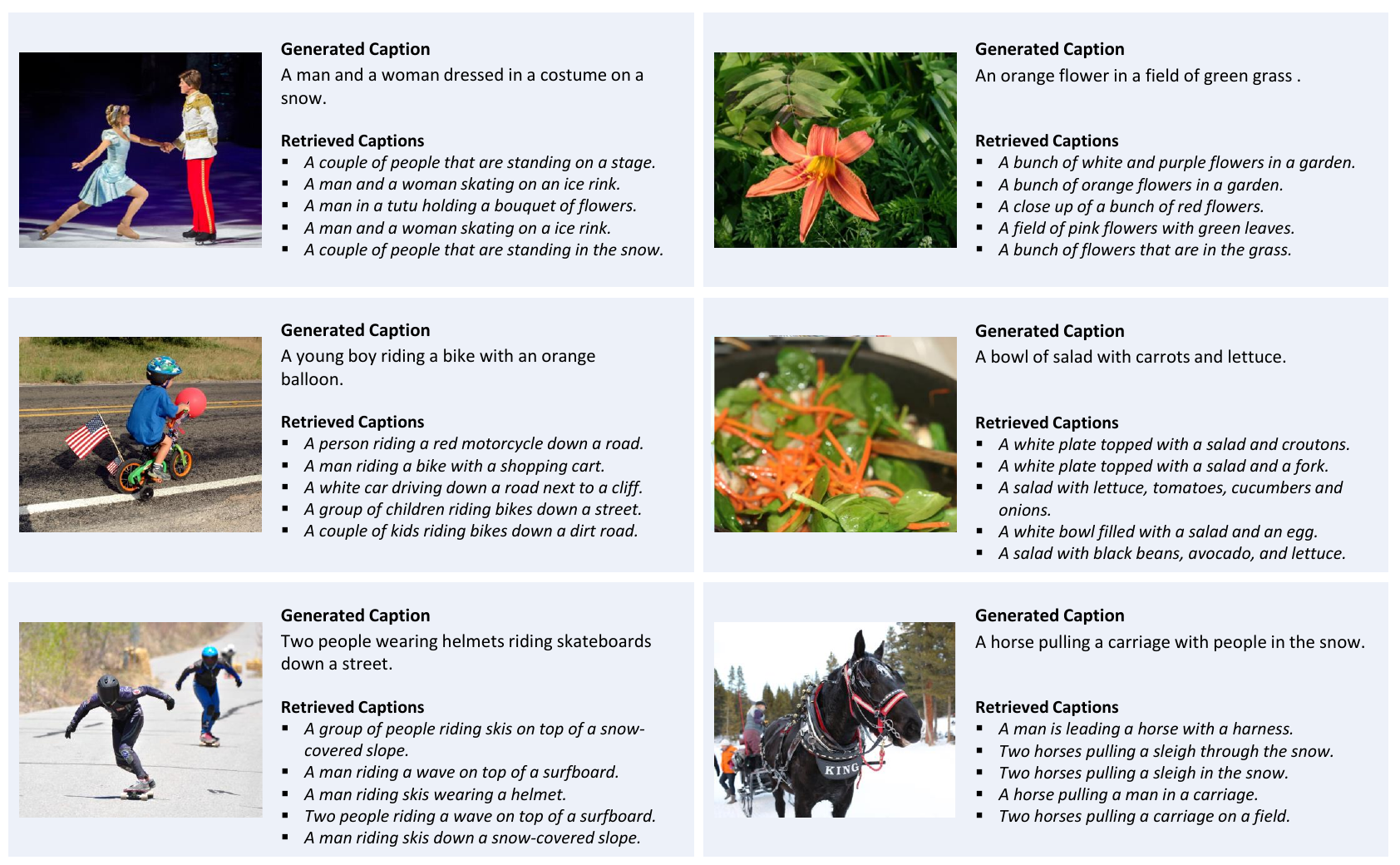}
    \caption{Generated captions on sample images from the COCO dataset, along with five retrieved captions.}
    \label{fig:qualitatives}
\end{figure*}

\tit{Results on nocaps}
In Table~\ref{tab:nocaps}, we extend our analysis on the nocaps dataset, using \oursself and \ourscross after finetuning with CIDEr optimization on the COCO dataset. \rev{Also in this case, we compare our results against a standard Transformer model and employ both versions of our retrieval index (\ie~the one containing COCO captions and the other composed of CC3M sentences predicted by the BLIP model).} The effectiveness of the proposed retrieval-augmented strategy is confirmed also in this setting, with an improvement of 1.5 and 1.7 on the entire validation set, respectively for the self- and cross-attention model variants with CC3M index. The contribution of a larger retrieval index becomes more evident, especially on near-domain and out-of-domain image-text pairs, which contain visual concepts outside of the COCO dataset and thus can benefit from a larger and semantically richer set of retrievable items. \rev{In fact, the CIDEr score on out-of-domain images is equal to 69.5 for the \oursself model with CC3M index compared to 68.6 achieved by the same version of the model augmented with COCO retrieval index.}

\begin{figure}[t]
\centering
\includegraphics[width=\columnwidth]{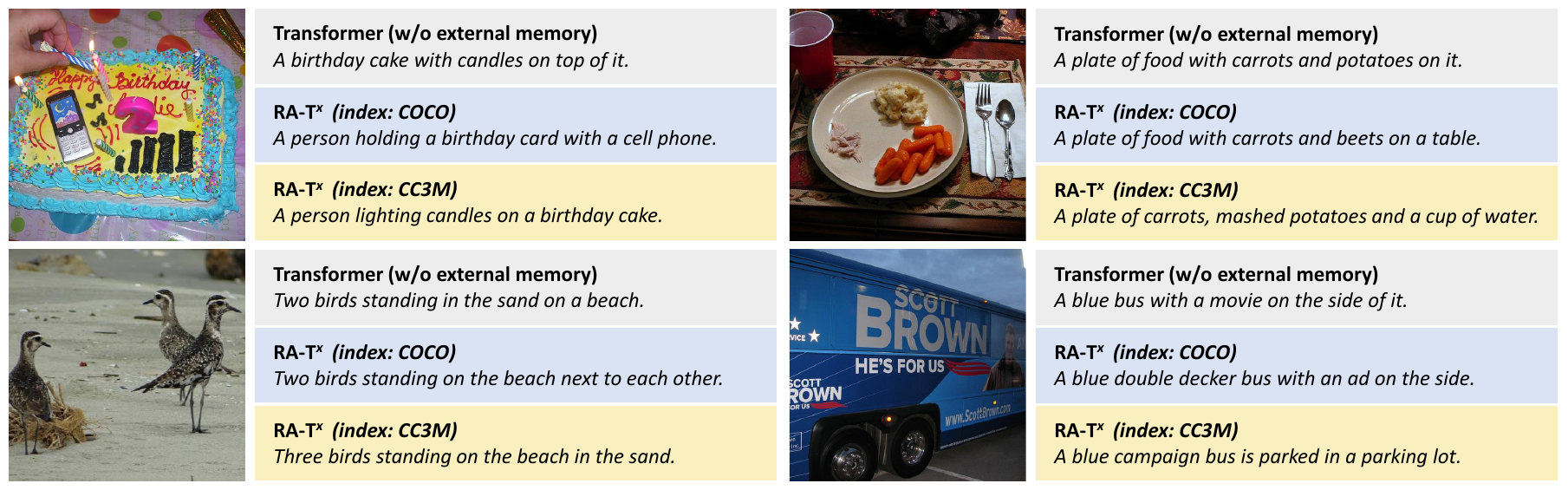}
\caption{Qualitative results on sample images from nocaps, comparing captions generated by our model with those generated by a standard Transformer without retrieval.}
\label{fig:nocaps}
\end{figure}

\tit{Qualitative Results}
Finally, in Fig.~\ref{fig:qualitatives} and~\ref{fig:nocaps} we report sample captions generated by our model on images respectively from COCO and nocaps. While in the former we show examples of captions retrieved from the external memory, in the latter we compare our results generated with both COCO and CC3M indexes to those generated by a Transformer-based model without the retrieval component. As it can be seen, retrieving captions related to the input image can help the language model generate more relevant and accurate captions by providing it with additional contextual information. For example, in the top-right example of Fig.~\ref{fig:qualitatives}, we can observe that the caption generated by our model has highly comparable content to the retrieved sentences (\ie~``orange flower'' and ``green grass''), providing evidence of the efficacy of our retrieval-based approach. When instead comparing our generated captions with those of a standard Transformer model without retrieval (Fig.~\ref{fig:nocaps}), we can observe that our results are generally more coherent with the visual content of input images and semantically richer, especially when using the retrieval index containing CC3M elements. \rev{For example, in the top-right image of Fig.~\ref{fig:nocaps}, the baseline model without external memory correctly recognizes the ``carrots'' and ``potataoes'' on a ``plate''. However, the caption predicted by our \ourscross model with CC3M index is more detailed and complete, also describing the presence of a ``cup of water'' and recognizing that the potatoes are ``mashed''. Similar observations can be made for the other examples, further confirming from a qualitative point of view the effectiveness of our retrieval-augmented solution for image captioning.}

\section{Conclusion}
In this paper, we have presented a novel framework for image captioning that is augmented with an external memory from which additional knowledge can be retrieved to help the generation process. The COCO and nocaps datasets were used to carry out experimental testing, which showed that adding retrieval capabilities to a captioning architecture can result in high-quality textual descriptions. This finding suggests that there is potential for additional research in this area.

%%
%% The acknowledgments section is defined using the "acks" environment
%% (and NOT an unnumbered section). This ensures the proper
%% identification of the section in the article metadata, and the
%% consistent spelling of the heading.
\begin{acks}
We acknowledge the CINECA award under the ISCRA initiative, for the availability of high-performance computing resources and support. This work has been conducted under a research grant co-funded by Leonardo S.p.A. and supported by the PNRR-M4C2 (PE00000013) project ``FAIR - Future Artificial Intelligence Research'', funded by the European Commission, and by the PRIN project ``CREATIVE: CRoss-modal understanding and gEnerATIon of Visual and tExtual content'' (CUP B87G22000460001), co-funded by the Italian Ministry of University and Research.
\end{acks}

%%
%% The next two lines define the bibliography style to be used, and
%% the bibliography file.
\bibliographystyle{ACM-Reference-Format}
\bibliography{bibliography}

\end{document}